\def\year{2022}\relax
\documentclass[letterpaper]{article} 
\usepackage{aaai22}  
\usepackage{times}  
\usepackage{helvet}  
\usepackage{courier}  
\usepackage[hyphens]{url}  
\usepackage{graphicx} 
\usepackage{natbib}  
\usepackage{caption} 
\DeclareCaptionStyle{ruled}{labelfont=normalfont,labelsep=colon,strut=off} 
\usepackage{algorithm}
\usepackage{algorithmic}
\usepackage{amsmath}
\usepackage{mathtools}
\usepackage{booktabs}
\usepackage{multirow}
\usepackage{subcaption}
\usepackage{newfloat}
\usepackage{listings}
\floatstyle{ruled}
\newfloat{listing}{tb}{lst}{}
\floatname{listing}{Listing}
\title{Compare Where It Matters: Using Layer-Wise Regularization To \\ Improve Federated Learning on Heterogeneous Data}
\author{
    Ha Min Son,\textsuperscript{\rm 1}
    Moon Hyun Kim,\textsuperscript{\rm 1}
    Tai-Myoung Chung,\textsuperscript{\rm 1}

}
\affiliations{
    \textsuperscript{\rm 1}Sungkyunkwan University\\

    Seoul, Republic of Korea\\

    sonhamin3@gmail.com, mhkim@skku.edu, tmchung@skku.edu
}

\begin{document}

\maketitle

\begin{abstract}
Federated Learning is a widely adopted method to train neural networks over distributed data. 
One main limitation is the performance degradation that occurs when data is heterogeneously distributed.
While many works have attempted to address this problem, these methods under-perform because they are founded on a limited understanding of neural networks.
In this work, we verify that only certain important layers in a neural network require regularization for effective training. We additionally verify that Centered Kernel Alignment (CKA) most accurately calculates similarity between layers of neural networks trained on different data. By applying CKA-based regularization to important layers during training, we significantly improve performance in heterogeneous settings.
We present FedCKA: a simple framework that out-performs previous state-of-the-art methods on various deep learning tasks while also improving efficiency and scalability.
\end{abstract}

\section{Introduction}
\noindent The success of deep learning in a plethora of fields has led to a countless number of research conducted to leverage its strengths \cite{lecun_deeplearning}. 
One main outcome resulting from this success is the mass collection of data \cite{terrence_dl}. As the collection of data increases at a rate much faster than that of the computing performance and storage capacity of consumer products, it is becoming progressively difficult to deploy trained state-of-the-art models within a reasonable budget.

Federated Learning (FL) \cite{mcmahan_fedavg} has been introduced as a method to train a neural network with massively distributed data. The most widely used and accepted approach for the training and aggregation process is FedAvg \cite{mcmahan_fedavg}. FedAvg is appealing for many reasons, such as negating the cost of collecting data into a centralized location, and effective parallelization across computing units \cite{verbraeken_distributed}. Thus, it has been applied to a wide range of researches, including a distributed learning framework on vehicular networks \cite{samarakoon_vehicular}, IoT devices \cite{yang_iot}, and 
even as a privacy-preserving method for medical records \cite{brisimi_medicalfl}.

One major issue with the application of FL is the performance degradation that occurs with heterogeneous data. This refers to settings in which data is not independent and identically distributed (non-IID) across clients. The drop in performance is seen to be caused by a disagreement in local optima. That is, because different clients train its copy of the neural network according to its individual local data, the resulting average can stray from the true optimum. Unfortunately, it is realistic to expect non-IID data in many real-world applications \cite{kairouz_advances, hsu_measuringNonIID}. In light of this, many works have attempted to address this problem by regularizing the entire model during the training process \cite{li_fedprox, karimireddy_scaffold, li_moon}. However, we argue that these works are based on a limited understanding of neural networks. 

In this work, we present FedCKA to address these limitations.
First, we show that regularizing the first two naturally similar layers are most important to improve performance in non-IID settings. Previous works had regularized each individual layers. Not only is this ineffective for training, it also limits scalability as the number of layers in a model increases. 
By regularizing only these important layers, performance improves beyond previous works. Efficiency and scalability is also improved, as we do not need to calculate regularization terms for every layer.
Second, we show that Centered Kernel Alignment (CKA) is most accurate when comparing the representational similarity between layers of neural networks. Previous works added a regularization term by comparing the representation of neural networks with simple inner products such as the l2-distance (FedProx) or cosine similarity (MOON). By using CKA to more accurately compare and regularize local updates, we improve performance; hence the name FedCKA. 
Our contributions are summarized as follows:
\begin{itemize}
\item We improve performance in heterogeneous settings. By building on the most up-to-date understanding of neural networks, we apply layer-wise regularization to only important layers.
\item We improve the efficiency and scalability of regularization. By regularizing only important layers, we exclusively show training times that are comparable to FedAvg.
\end{itemize}

\begin{figure*}[t]
\centering
\includegraphics[width=0.75\textwidth]{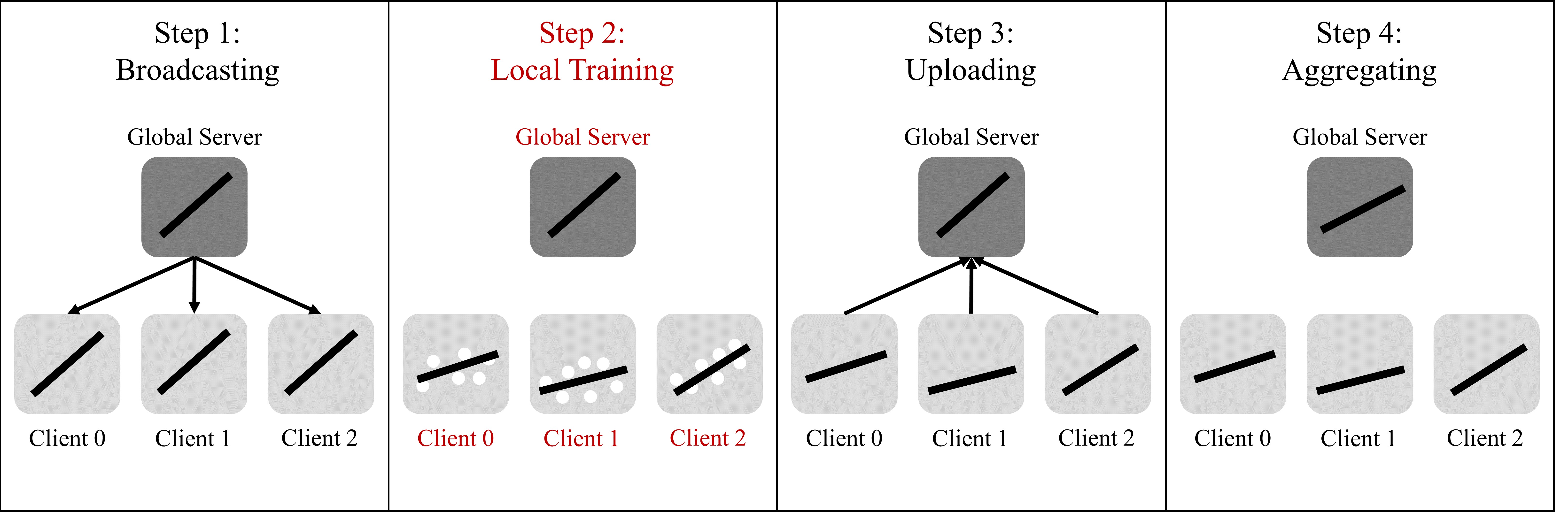} 
\caption{The typical steps of Federated Learning}
\label{fig:fedlearning_steps}
\end{figure*}

\section{Related Works}

\subsection{Layers in Neural Networks}
Understanding the function of layers in a neural network is an under-researched field of deep learning. It is, however, an important prerequisite for the application of layer-wise regularization. We build our work based on findings of two relevant papers. 

The first work \cite{zhang_layerscreated} showed that there are certain 'critical' layers that define a model's performance. In particular, when layers were re-initialized back to their original weights, 'critical' layers heavily decreased performance, while 'robust' layers had minimal impact. This work drew several relevant conclusions. First, the very first layer of neural networks is most sensitive to re-initialization. Second, robustness is not correlated with the l2-norm or l$\infty$-norm between initial weights and trained weights. 
Considering these conclusions, we understand that certain layers are not important in defining performance. Regularizing these non-important layers would be ineffective, and may even hurt performance. 

The second work \cite{kornblith_CKAsim} introduced Centered Kernel Alignment (CKA) as a metric for measuring the similarity between layers of neural networks. In particular, the work showed that metrics that calculate the similarity between representations of neural networks should be invariant to orthogonal transformations and isotropic scaling, while invertible to linear transformations. This work drew one very relevant conclusion. For neural networks trained on different datasets, early layers, but not late layers, learn similar representations.
Considering this conclusion, if we were to properly regularize neural networks trained on different datasets, we should focus on layers that are naturally similar, and not on those that are naturally different. 

\subsection{Federated Learning with Non-IID Data}

Federated Learning typically progresses with the repetition of four steps as shown in Figure \ref{fig:fedlearning_steps}. 
1) a centralized or de-centralized server broadcasts a model (the global model) to each of its clients. 2) Each client trains its copy of the model (the local model) with its local data. 3) The client uploads its trained model to the server. 4) The server aggregates the trained model into a single model and prepares it to be broadcast in the next round. These steps are repeated until convergence or other criteria are met. 

Works that improve performance on non-IID data generally falls into two categories. The first focuses on regularizing or modifying the client training process (step 2). The second focuses on modifying the aggregation process (step 4). Here, we focus on the former, as it is more closely related to our work. Namely, we focus on FedProx \cite{li_fedprox}, SCAFFOLD \cite{karimireddy_scaffold}, and MOON \cite{li_moon}, all of which add a regularization term to the default FedAvg \cite{mcmahan_fedavg} training process.

FedAvg was the first work to introduce Federated Learning. Each client trains a model using a gradient descent loss function, and the server averages the trained model based on the number of data samples each client holds. 
However, due to the performance degradation in non-IID settings, many works have added a regularization term to the default FedAvg training process. The objective of these methods is to decrease the disagreement in local optima by limiting local updates that stray too far from the global model.
FedProx adds a proximal regularization term that calculates the l2-distance between the local and global model. 
SCAFFOLD adds a control variate regularization term that induces variance reduction on local updates based on the updates of other clients.
Most recent and most similar to our work is MOON. MOON adds a contrastive regularization term that calculates the cosine similarity between the MLP projections of the local and global model. The work takes inspiration from contrastive learning, in particular, SimCLR \cite{chen_simclr}. The intuition is that the global model is less biased than local models, thus local updates should be more similar to the global model than past local models. One difference to note is that while contrastive learning trains a model using the projections of one model on many different images (i.e. one model, different data), MOON regularizes a model using the projections of different models on the same images (i.e. three models, same data).

Overall, these works add a regularization term by comparing all layers of the neural network. However, we argue that only important layers should be regularized. Late layers are naturally dissimilar when trained on different datasets. Regularizing a model based on these naturally dissimilar late layers would be ineffective. Rather, it may be beneficial to focus only on the earlier layers of the model.

\section{FedCKA}

\subsection{Regularizing Naturally Similar Layers}
\begin{figure}[t]
\centering
    \includegraphics[width=0.9\columnwidth]{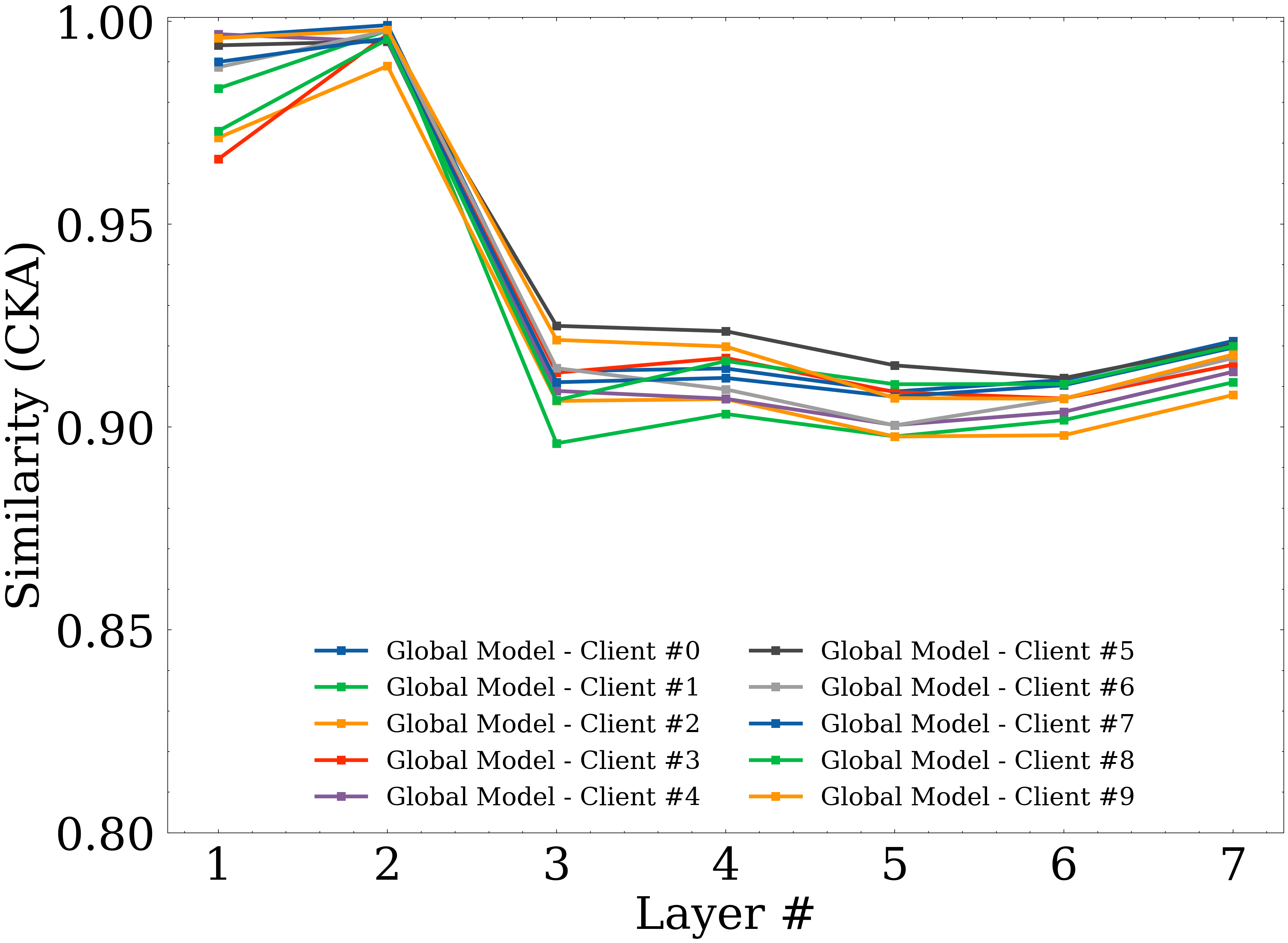} 
\caption{CKA similarity comparison between each client and the global model (Refer to \textit{Experimental Setup} for more information on setup)}
\label{fig_ckacomp}
\end{figure}

FedCKA is designed on the principle that naturally similar, but not naturally dissimilar, layers should be regularized. This is based on the premise that early layers, but not late layers, develop similar representations when trained on different datasets \cite{kornblith_CKAsim}. We verify this in a Federated Learning environment. Using a small Convolutional Neural Network, we trained 10 clients for 20 communications rounds on independently and identically distributed (IID) subsets of the CIFAR-10 \cite{cifar10_100} dataset. After training, we viewed the similarity between each layer of the local and global models, calculated by the Centered Kernel Alignment \cite{kornblith_CKAsim} on the CIFAR-10 test set. The similarity of each layer between local and global models are shown in Figure \ref{fig_ckacomp}. We verify that early layers, but not late layers, develop similar representations even in the most optimal Federated Learning setting, where the distribution across data between clients are IID. 

The objective of regularizing local updates is to penalize updates that stray from the global model. However, late layers are naturally dissimilar even in optimal Federated Learning settings. If this is the case, regularizing these late layers would penalize updates that may have been beneficial to training. Thus, FedCKA regularizes only the first two naturally similar layers. For convolutional neural networks without residual blocks, the first two naturally similar layers are the two layers closest to the input. For ResNets \cite{he_resnet}, it is the initial convolutional layer and first post-residual block. As also mentioned in \citet{kornblith_CKAsim}, post-residual layers, but not layers within residuals, develop similar representations. This is unique to previous works, which had regularized local updates based on all layers. This also allows FedCKA to be much more scalable than other methods. The computational overhead for previous works increases rapidly in proportion to the number of parameters, because all layers are regularized. FedCKA keeps the overhead nearly constant, as we regularize only two layers close to the input.

\subsection{Measuring Layer-wise Similarity}
FedCKA is designed to regularize dissimilar updates in layers that should naturally be similar.
However, there is currently no standard for measuring the similarity of layers between neural networks. While there are classical methods of applying univariate or multivariate analysis for comparing matrices, these methods are not suitable for comparing the similarity of layers and representations of different neural networks \cite{kornblith_CKAsim}. As for norms, \citet{zhang_layerscreated} concluded that a layer's robustness to re-initialization is not correlated with the l2-norm or l$\infty$-norm. This suggests that using these norms to regularize dissimilar updates, as in previous works, may be inaccurate.

\citet{kornblith_CKAsim} concluded that similarity metrics for comparing the representation of different neural networks should be invariant to orthogonal transformations and isotropic scaling, while invertible to linear transformation. The work introduced Centered Kernel Alignment (CKA), and showed that the metric is most consistent in measuring the similarity between representation of neural networks. Thus, FedCKA regularizes local updates using the CKA metric as a similarity measure.

\subsection{Modifications to FedAvg}
\begin{figure}[t]
\centering
    \includegraphics[width=0.9\columnwidth]{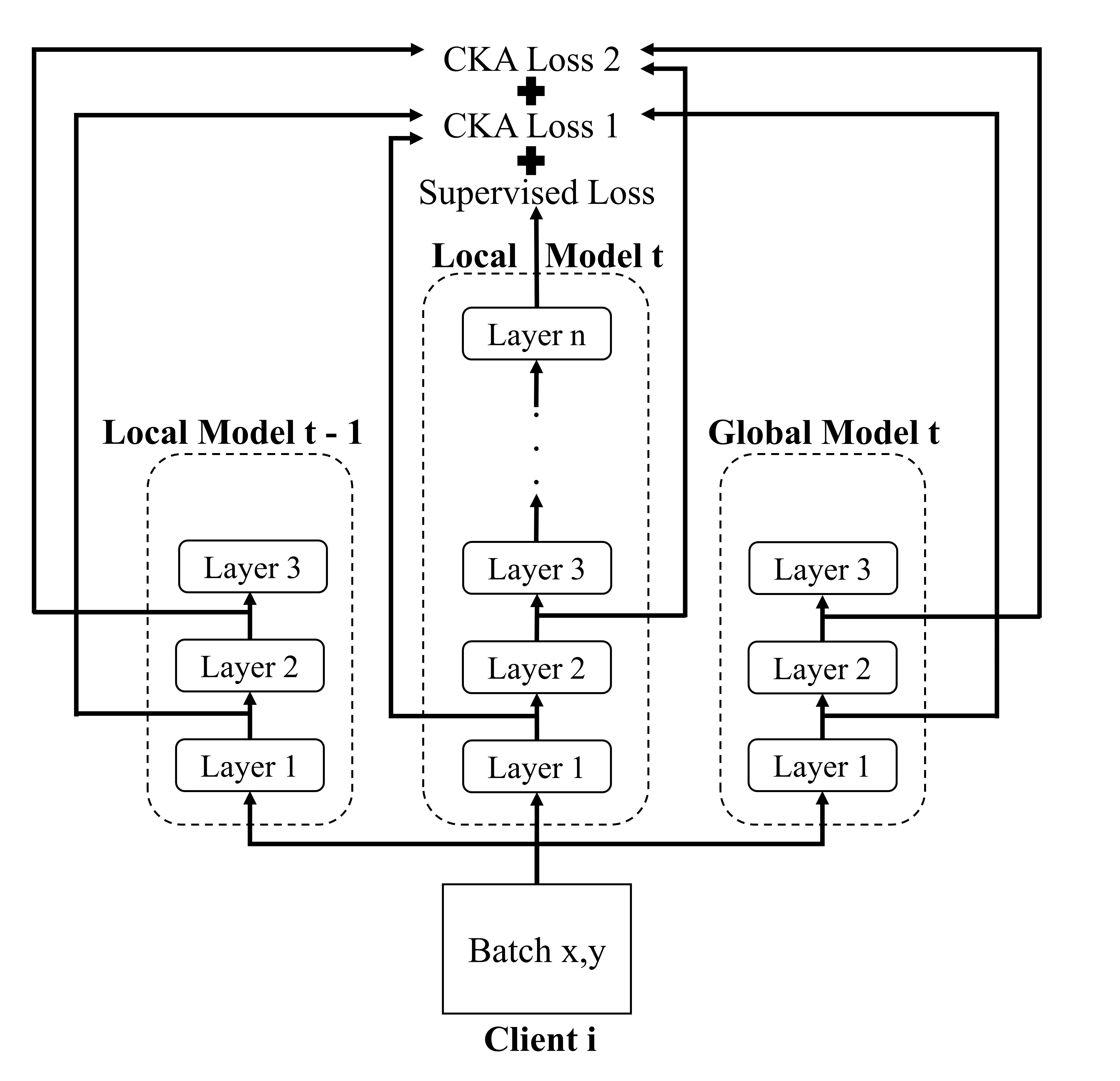}
\caption{Training Process of FedCKA}
\label{fig:training}
\end{figure}

FedCKA adds a regularization term to the local training process of the default FedAvg algorithm, keeping the entire framework simple. Alg \ref{alg:fedcka} and Fig \ref{fig:training} shows the FedCKA framework in algorithm and figure form, respectively.  More formally, we add $\ell_{cka}$ as a regularization term to the FedAvg training algorithm. The local loss function is as shown in Eq \ref{local_obj_eq}.
\begin{equation}
    \label{local_obj_eq}
     \ell = 
     \ell_{sup}(w^{t}_{l_i}; D^{i}) + \mu \ell_{cka}(w^{t}_{l_i}; w^{t}_{g} ;w^{t-1}_{l_i}; D^{i})
\end{equation}

\noindent Here, $\ell_{sup}$ is the cross entropy loss, $\mu$ is a hyper-parameter to control the strength of the regularization term, $\ell_{cka}$, in proportion to $\ell_{sup}$. $\ell_{cka}$ is shown in more detail in Eq \ref{l_cka_eq}.

The formula of $\ell_{cka}$ is a slight modification to the contrastive loss that is used in SimCLR \cite{chen_simclr}. There are four main differences. 
First, SimCLR uses the representations of one model on different samples in a batch to calculate contrastive loss. FedCKA uses the representation of three models on the same samples in a batch to calculate $\ell_{cka}$. $a^{t}_{l_{i}}$, $a^{t-1}_{l_{i}}$, and $a^{t}_{g}$ are the representations of client $i$'s current local model, client $i$'s previous round local model, and the current global model, respectively.
Second, SimCLR uses the temperature parameter $\tau$ to increase performance on difficult samples. FedCKA excludes $\tau$, as it was not seen to help performance. 
Third, SimCLR uses cosine similarity to measure the similarity between the representations of difference datasets. FedCKA uses CKA as its measure of similarity. 
Fourth, SimCLR calculates contrastive loss once per batch, using the representations of the projection head. FedCKA use calculates $\ell_{cka}$ $M$ times per batch, using the representations of the first $M$ naturally similar layers, indexed by $n$, and averages the loss based on the number of layers to regularize. $M$ is set to two by default unless otherwise stated.

\begin{algorithm}[tb]
\caption{FedCKA}
\label{alg:fedcka}
\textbf{Input}: number of communication rounds R, number of clients C, number of local epochs E, loss weighting variable $\mu$, learning rate $\eta$ \\
\textbf{Output}: The trained model $w$
\begin{algorithmic}[1] 
\STATE Initialize $w^{0}_{g}$
\FOR{each round $r \ \in \ [0, R-1]$}
    \FOR{each client $i \ \in \ [0, C]$}
        \STATE $w^{t}_{l_{i}} \leftarrow$  LocalUpdate($w^{t}_{g}$)
    \ENDFOR
    \STATE $W^{t}_l \leftarrow [w^{t}_{l_{0}}$, $w^{t}_{l_{1}}$, ... , $w^{t}_{l_{C-1}}]$
    \STATE $w^{t}_{g} \leftarrow$  WeightedAvg($W^{t}_l$)
\ENDFOR
\STATE \textbf{return} $w^{t}_{g}$ \\\

\STATE \textbf{LocalUpdate}($w^{t}_{g}$):
\STATE $w^{t}_{l_{i}} \leftarrow w^{t}_{g}$
    \FOR{each epoch $e \ \in \ [0, E-1]$}
        \FOR{each batch $b \ \in \ D^{i}$}
            \STATE $\ell_{sup} \leftarrow CrossEntropyLoss(w^{t}_{l_{i}}; b)$
            \STATE $\ell_{cka} \leftarrow CKALoss(w^{t}_{l_{i}}; w^{t-1}_{l_{i}}; w^{t}_{l_{i}}; b)$
            \STATE $\ell \leftarrow \ell_{sup} + \mu \ell_{cka}$
            \STATE $w^{t}_{l_{i}} \leftarrow w^{t}_{l_{i}} - \eta \nabla \ell$
        \ENDFOR
    \ENDFOR
\STATE \textbf{return} solution  \\\


\STATE \textbf{WeightedAvg}($W^{t}_{l}$):
\STATE Initialize $w^{t}_{g}$ 
\FOR{each client $i \ \in \ [0, C]$}
    \STATE  $w^{t}_{g} \mathrel{+}= \frac{\left | D^{i} \right |}{\left | D \right |} W^{t}_{l_{i}}$
\ENDFOR
\STATE \textbf{return} $w^{t}_{g}$ 
\end{algorithmic}
\end{algorithm}

\begin{equation}
    \label{l_cka_eq}
    \begin{multlined}
        \ell_{cka} = \\
        \frac{1}{M}
        \sum_{n=1}^{M}
        -\log
        \left (  
        \frac{
        e^{(CKA(a^{t}_{l_{i_{n}}}, a^{t}_{g_{n}})}}
        {e^{(CKA(a^{t}_{l_{i_{n}}}, a^{t}_{g_{n}}))} + e^{(CKA(a^{t}_{l_{i_{n}}}, a^{t-1}_{l_{i_{n}}}))}
        }
        \right )
    \end{multlined}
\end{equation}

\noindent As per \citet{kornblith_CKAsim}, CKA is shown in Eq \ref{CKA_eq}. Here, the $i^{th}$ eigenvalue of $XX^{T}$ is $\lambda^{i}_{X}$. 

\begin{equation}
    \label{CKA_eq}
    \begin{multlined}
        CKA(XX^{T}, YY^{T}) =  
        \frac{\left \| Y^{T}X \right \|^{2}_{F}} {\left \| X^{T}X \right \|_{F} \left \| Y^{T}Y \right \|_{F}}
        \\ = \frac{ \sum_{i=1}^{p1} \sum_{j=1}^{p2} 
        \lambda^{i}_{X} \lambda^{i}_{Y} 
        \left \langle \textbf{u}^{i}_{X}, \textbf{u}^{j}_{Y} \right \rangle ^{2}
        }
        {
        \sqrt{\sum_{i=1}^{p1}(\lambda^{i}_{X})^{2}}  
        \sqrt{\sum_{j=1}^{p2}(\lambda^{j}_{Y})^{2}}  
        }        
    \end{multlined}
\end{equation}

\noindent While \citet{kornblith_CKAsim} also presented a method to use kernels with CKA, we use the linear variant, as it is more computationally efficient, while having minimal impact on accuracy.

\begin{figure*}[t]
\centering
     \begin{subfigure}[t]{0.3\textwidth}
         \centering
         \includegraphics[width=\textwidth]{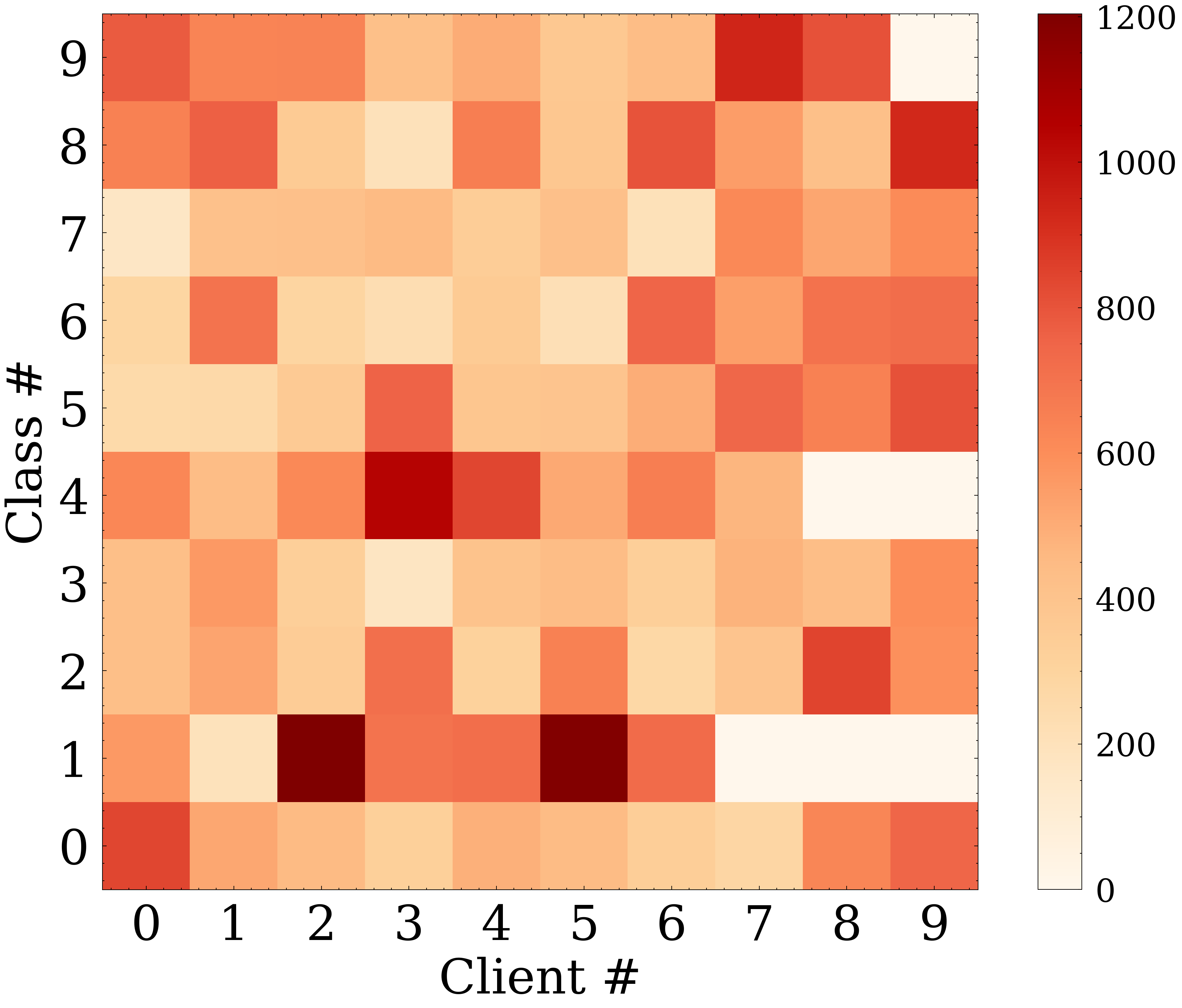}
         \caption{$\alpha = 5.0$}
         \label{fig:alpha50}
     \end{subfigure}
     \hfill
     \begin{subfigure}[t]{0.3\textwidth}
         \centering
         \includegraphics[width=\textwidth]{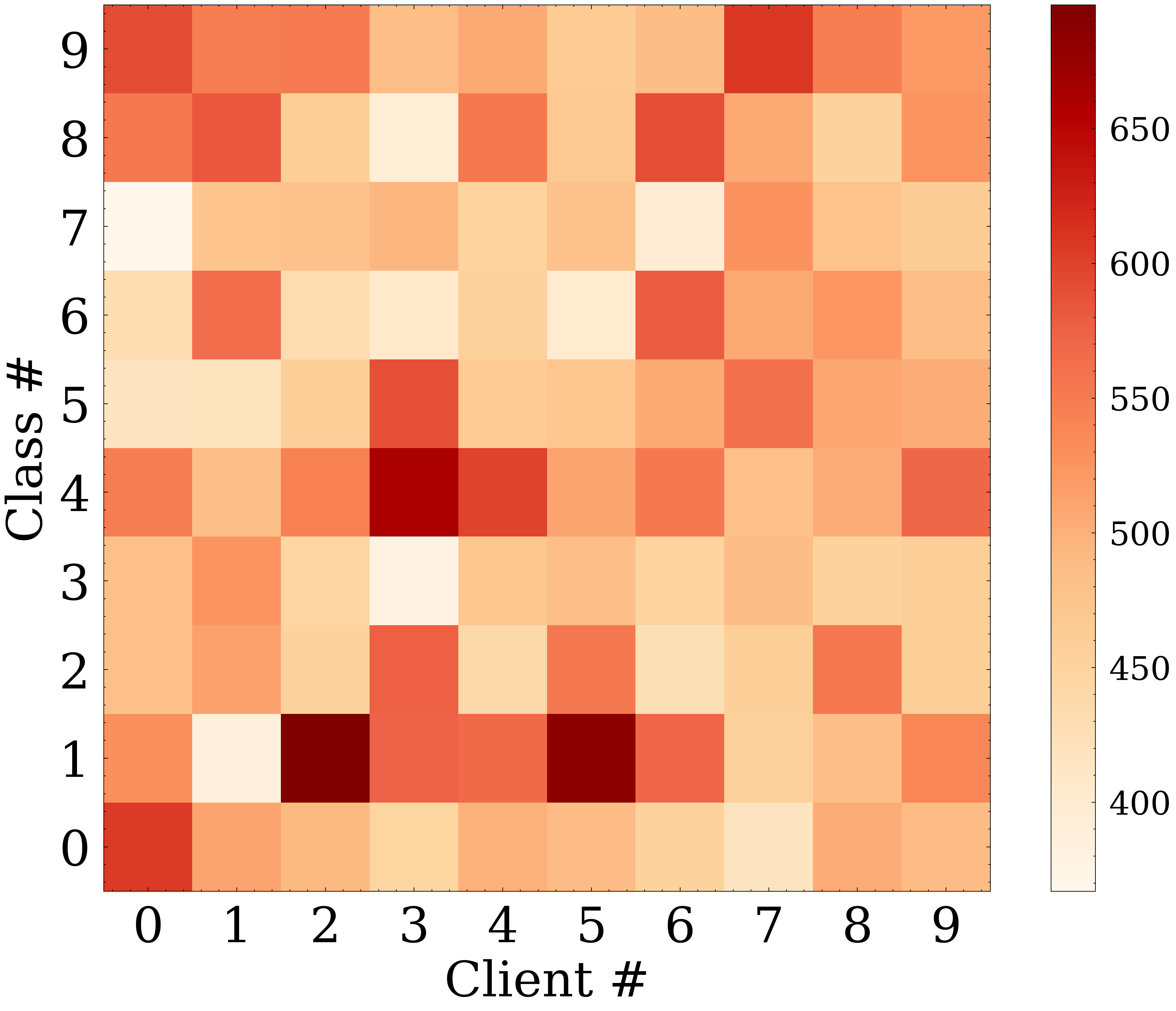}
         \caption{$\alpha = 0.5$}
         \label{fig:alpha05}
     \end{subfigure}
     \hfill
     \begin{subfigure}[t]{0.3\textwidth}
         \centering
         \includegraphics[width=\textwidth]{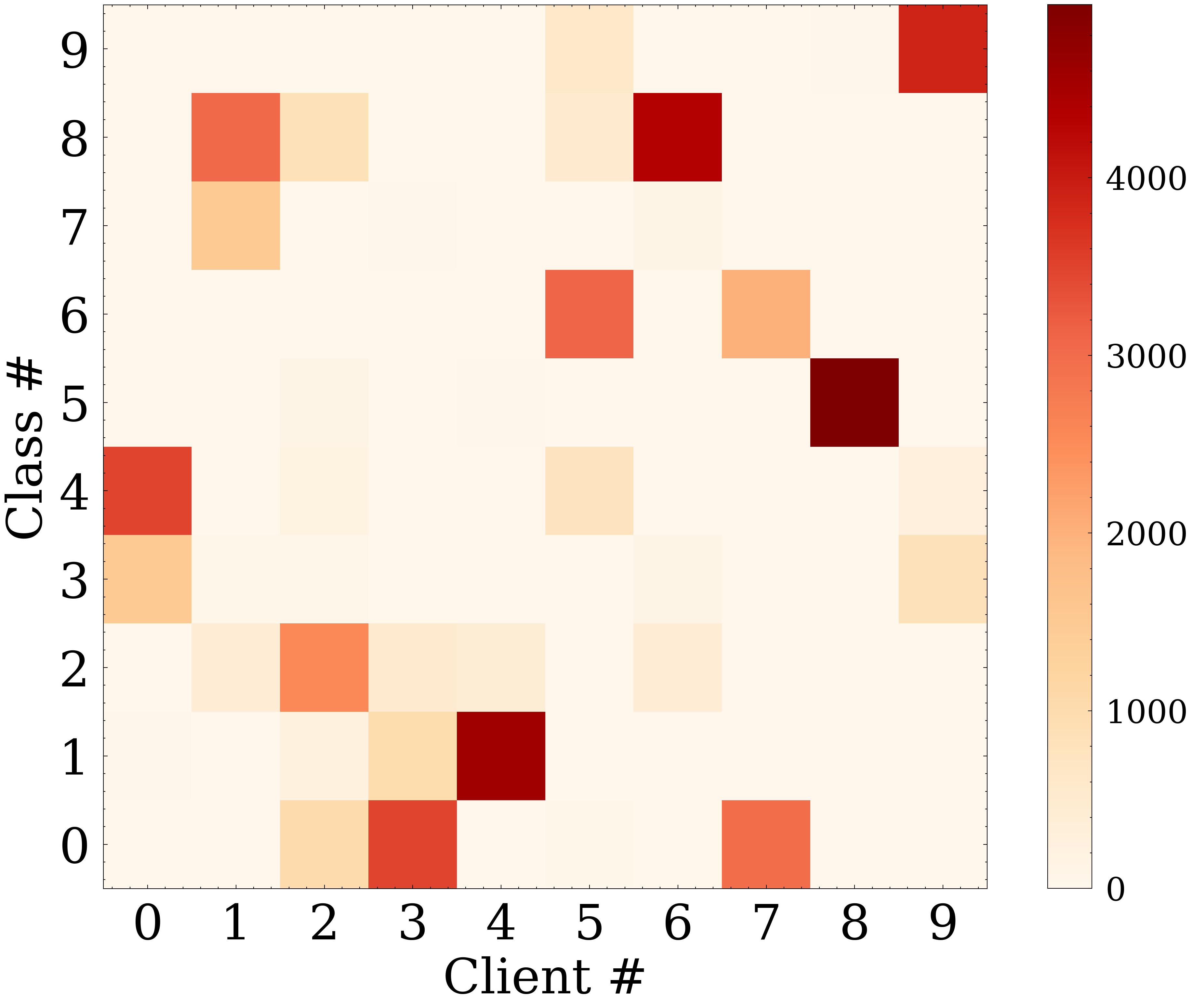}
         \caption{$\alpha = 0.1$}
         \label{fig:alpha01}
     \end{subfigure}
\caption{Distribution of the CIFAR-10 dataset across 10 clients according to the Dirichlet distribution. The x-axis shows the index of the client, and the y-axis shows the index of the class (label). As the parameter $\alpha$ approaches 0, there the heterogeneity of class distribution increases.}
\label{fig:heatmaps}
\end{figure*}

\section{Experimental Results and Analysis}

\subsection{Experiment Setup}
We compare FedCKA with the current state-of-the-art, MOON \cite{li_moon}, as well as FedAvg \cite{mcmahan_fedavg}, FedProx \cite{li_fedprox}, and SCAFFOLD \cite{karimireddy_scaffold}.
We purposefully use a similar experimental setup to MOON, both because it is the most recent work, and also reports the highest performance. In particular, the CIFAR-10, CIFAR-100 \cite{cifar10_100}, and Tiny ImageNet \cite{tiny_imagenet} datasets are used to test the performance of all methods. 

For CIFAR-10, we use a small Convolutional Neural Network. Two 5x5 convolutional layers are the base encoder, with 16 and 32 channels respectively, and two 2x2 max-pooling layers following each convolutional layer. A projection head of four fully connected layers follow the encoder, with 120, 84, 84, and 256 neurons. The final layer is the output layer with the number of classes. Although FedCKA and other works can perform without this projection head, we include it because MOON shows a high discrepancy in performance without it. 
For CIFAR-100 and Tiny ImageNet, we use ResNet-50 \cite{he_resnet}. We also add the projection head before the output layer, as per MOON.

We use the cross entropy loss, and SGD as our optimizer with a learning rate of 0.1, momentum of 0.9, and weight decay of 0.00001. Local epochs are set to 10. These are also the parameters used in MOON. Some small changes we made were with the batch size and communication rounds. We use a constant 128 for the batch size, and train for 100 communication rounds on CIFAR-10, and 40 communication rounds on CIFAR-100, and 20 communication rounds on Tiny ImageNet. We use a lower number of communication rounds for the latter two datasets, because the ResNet-50 model over-fit quite quickly.

As with many previous works, we use the Dirichlet distribution to simulate heterogeneous settings \cite{distribution_ex1, distribution_ex2, li_moon}. The $\alpha$ parameter controls the strength of heterogeneity, with $\alpha = 0$ being most heterogeneous, and $\alpha = \infty$ being non-heterogeneous. We report results for $\alpha \in [5.0, 0.5, 0.1]$, similar to MOON. Figure \ref{fig:heatmaps} shows the distribution of data across clients on the CIFAR-10 dataset with the different $\alpha$.
All experiments were conducted using the PyTorch \cite{pytorch} library on a single GTX Titan V and four Intel Xeon Gold 5115 processors.

\subsection{Accuracy}
FedCKA adds a hyperparameter $\mu$ to control the strength of $\ell_{cka}$. We tune $\mu$ from [3, 5, 10], and report the best results. MOON and FedProx also have a $\mu$ term. We also tune the hyperparameter $\mu$ with these methods. For MOON, we tune $\mu$ from [0.1, 1, 5, 10] and for FedProx, we tune $\mu$ from [0.001, 0.01, 0.1, 1], as used in each work. In addition, for MOON, we use $\tau = 0.5$ as reported in their work.

\begingroup
\setlength{\tabcolsep}{7pt} 
\renewcommand{\arraystretch}{1} 
\begin{table}[]
\centering
\begin{tabular}{@{}c|c|c|c@{}}
\toprule
Method          & CIFAR-10 & CIFAR-100 & \begin{tabular}[c]{@{}c@{}}Tiny \\ ImageNet\end{tabular} \\ \midrule
FedAvg          &           64.37\%  &          37.41\%  &          19.49\%  \\
FedProx         &           64.58\%  &          37.81\%  &          20.93\%  \\
SCAFFOLD        &           64.33\%  &          39.16\%  &          21.18\%  \\
MOON            &           65.25\%  &          38.37\%  &          21.29\%  \\
\textbf{FedCKA} &   \textbf{67.86\%} &  \textbf{40.07\%} &  \textbf{21.46\%}  \\ \bottomrule
\end{tabular}
\caption{Accuracy across Datasets ($\alpha = 5.0$)}
\label{tab:acc_datasets}
\end{table}
\endgroup

Table \ref{tab:acc_datasets} shows the performance across CIFAR-10, CIFAR-100, and Tiny ImageNet with $\alpha = 5.0$. For FedProx, MOON, and FedCKA, we report performance with the best $\mu$. For FedCKA, the best $\mu$ is 3, 10, and 3 for CIFAR-10, CIFAR-100, and Tiny ImageNet, respectively. For MOON, the best $\mu$ is 10, 5, and 0.1.  For FedProx, the best $\mu$ is 0.001, 0.1, and 0.1. Table \ref{tab:acc_alpha} shows the performance across increasing heterogeneity on the CIFAR-10 dataset with $\alpha \ \in \ [5.0, 0.5, 0.1]$. For FedCKA, the best $\mu$ is 5, 3, and 3 for each $\alpha \ \in \ [5.0, 0.5, 0.1]$, respectively. For MOON, the best $\mu$ is 0.1, 10, and 10. For FedProx, the best $\mu$ is 0.001, 0.1, and 0.001.

\begingroup
\setlength{\tabcolsep}{10pt} 
\renewcommand{\arraystretch}{1} 
\begin{table}[]
\centering
\begin{tabular}{@{}c|c|c|l@{}}
\toprule
\multirow{2}{*}{Method} & \multicolumn{1}{c|}{\multirow{2}{*}{$\alpha$ = 5.0}} & \multicolumn{1}{c|}{\multirow{2}{*}{$\alpha$ = 0.5}} & 
\multicolumn{1}{c}{\multirow{2}{*}{$\alpha$ = 0.1}} \\
& \multicolumn{1}{c|}{} & \multicolumn{1}{c|}{} & \multicolumn{1}{c}{}       \\ \midrule
FedAvg          &           64.37\%  &          59.81\%  &          50.43\%  \\
FedProx         &           64.58\%  &          59.98\%  &          51.07\%  \\
SCAFFOLD        &           64.33\%  &          59.47\%  &          40.53\%  \\
MOON            &           65.25\%  &          60.65\%  &          51.63\%  \\
\textbf{FedCKA} &   \textbf{67.86\%}  & \textbf{61.13\%} &  \textbf{52.35\%} \\ \bottomrule
\end{tabular}
\caption{Accuracy across $\alpha \ \in \ [5.0, 0.5, 0.1]$ (CIFAR-10)}
\label{tab:acc_alpha}
\end{table}
\endgroup

We observe that FedCKA consistently outperforms previous methods across different datasets and across different $\alpha$. FedCKA improves performance in heterogeneous settings owing to regularizing layers that are naturally similar, and not layers that are naturally dissimilar. It is also interesting to see that FedCKA performs better by a larger margin when $\alpha$ is larger. This is likely because the global model is less biased as data distribution approaches IID settings, thus can more effectively regularize updates. However, we also observe that other works consistently improve performance, albeit by a smaller margin than FedCKA. FedProx and SCAFFOLD improve performance likely owing to their inclusion of naturally similar layers in regularization. The performance gain is lower, as they also include naturally dissimilar layers in regularization. MOON improves performance compared to FedProx and SCAFFOLD likely owing to their use of a contrastive loss. That is, MOON shows that neural networks should be trained to be more similar to the global model \textit{than past local model}, rather than only be blindly similar the global model. By regularizing naturally similar layers using a contrastive loss based on CKA, FedCKA outperforms all methods.

Note that across most methods and settings, there are discrepancies to the accuracy reported by MOON \cite{li_moon}. In particular, MOON reports higher accuracy across all methods although model architecture are similar, if not equivalent. We suspect that data augmentation was used to increase accuracy. We could not test these settings, as MOON did not record their parameters. We thus report result without data augmentation techniques.

\subsection{Regularizing Only Important Layers}
\begin{figure}[t]
\centering
\includegraphics[width=0.9\columnwidth]{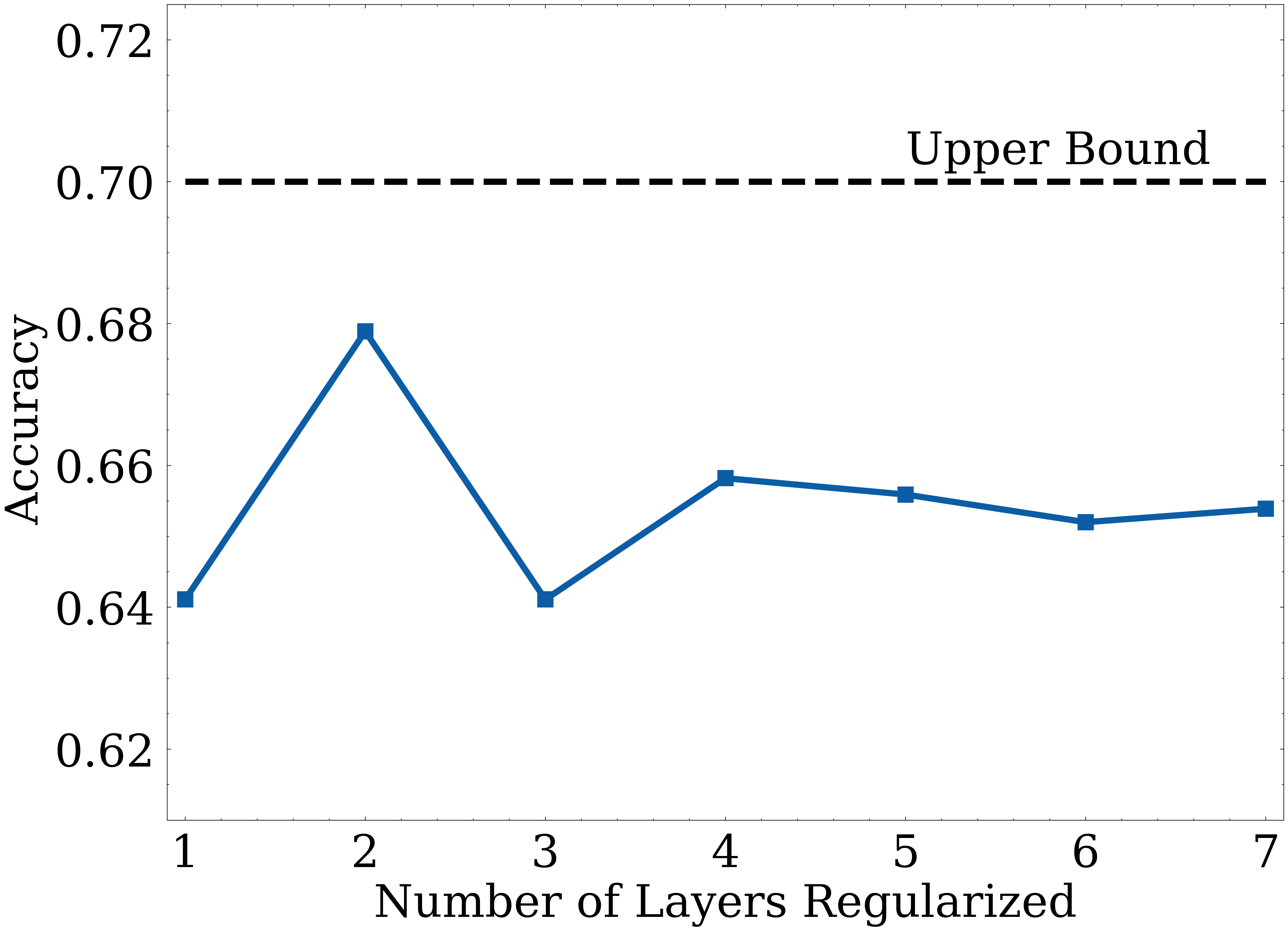} 
\caption{Accuracy with respect to the number of layers regularized on CIFAR-10 and $\alpha = 5.0$}
\label{fig:layer_acc}
\end{figure}

We study the effects of regularizing different number of layers. Using the CIFAR-10 dataset with $\alpha = 5.0$, we change the number of layers to regularize through $\ell_{cka}$. Formally, we change $M$ in Eq \ref{l_cka_eq} by scaling  $M \ \in \ [1, 2, 3, 4, 5, 6, 7]$, and report the accuracy in Figure \ref{fig:layer_acc}. 
Accuracy is highest when only the first two layers are regularized. This verifies our claim that only naturally similar, but not naturally dissimilar layers should be regularized (Figure \ref{fig_ckacomp}).
In addition, note the dotted line representing the upper bound for Federated Learning. When the same model is trained on a centralized server with the whole CIFAR-10 dataset, accuracy is 70\%. FedCKA with regularization on the first two naturally similar layers nearly reaches this upper bound.

\begingroup
\setlength{\tabcolsep}{7pt} 
\renewcommand{\arraystretch}{1} 
\begin{table}[]
\centering
\begin{tabular}{@{}c|c|c@{}}
\toprule
Similarity Metric & Accuracy & \begin{tabular}[c]{@{}c@{}}Training \\ Duration (s)\end{tabular}  \\ \midrule
None (FedAvg)    &          64.37\%  &          54.82  \\
Frobenius Norm        &          65.54\%  &          64.73  \\
Vectorized Cosine           &          66.67\%  &          65.75  \\
Kernel CKA       &          67.93\%  &          122.41  \\
Linear CKA       &          67.86\% &           104.17     \\ \bottomrule
\end{tabular}
\caption{Accuracy and training duration with FedCKA with different similarity metrics (CIFAR-10)}
\label{tab:different_sims}
\end{table}
\endgroup

\subsection{Using the Best Similarity Metric}
We study the effects of regularizing the first two naturally similar layers with different similarity metrics. Using the CIFAR-10 dataset with $\alpha = 5.0$, we change the similarity metric to regularize through $\ell_{cka}$. Formally, we change $CKA(a_{1}, a_{2})$ in Eq \ref{l_cka_eq} to three other similarity metrics. First, the kernel CKA, introduced in \citet{kornblith_CKAsim} ($CKA_{k}(a_{1}, a_{2})$). Second, the squared Frobenius norm ($\left \| a_{1} - a_{2}  \right \|_{F}^{2}$). Third, the vectorized cosine similarity ($\left \| vec(a_{1}) \right \| \left \| vec(a_{2}) \right \| \cos \theta $). We compare the results with these different metrics as well as the baseline, FedAvg. The results are shown in Table \ref{tab:different_sims}. 

We observe that performance is highest when CKA is used. This is likely owing to the accuracy of measuring similarity. Only truly dissimilar updates are penalized, thus improving performance. In addition, while kernel CKA slightly outperforms linear CKA, considering the computational overhead, we opt to use linear CKA.
We also observe that the squared Frobenius norm and vectorized cosine similarity decrease performance only slightly. These methods outperform most previous works. This verifies that while it is important to use an accurate similarity measure, it is more important to focus on regularizing naturally similar layers. 

\begingroup
\setlength{\tabcolsep}{4pt} 
\renewcommand{\arraystretch}{1} 
\begin{table}[]
\centering
\begin{tabular}{@{}l|cc|cc@{}}
\toprule
Method          & 7 Layers & \begin{tabular}[c]{@{}c@{}}Time\\Extended\end{tabular} & 50 Layers & \begin{tabular}[c]{@{}c@{}}Time\\Extended\end{tabular} \\ \midrule
FedAvg          & 54.82                                                           & -             & 638.79                                                          & -             \\
SCAFFOLD        & 57.19                                                           & 2.37          & 967.04                                                          & 328.25        \\
FedProx         & 57.20                                                           & 2.38          & 862.12                                                          & 223.33        \\
MOON            & 97.58                                                           & 42.76         & 1689.28                                                         & 1050.49       \\
\textbf{FedCKA} & 104.17                                                          & \textbf{49.35}         & 750.97                                                          & \textbf{112.18}        \\ \bottomrule
\end{tabular}
\caption{Average Training Duration Per Communication Round (in seconds)}
\label{tab:times}
\end{table}
\endgroup

\subsection{Efficiency and Scalability}
Efficient and scalable local training is an important engineering principle of Federated Learning. That is, for Federated Learning to be applied to real-world applications, we must assume that clients have limited computing resources. Thus, we analyze the local training time of all methods, as shown in Table \ref{tab:times}.
Note that FedAvg is the lower bound for training time, since all other methods add a regularization term. 

For a 7-layer CNN trained on CIFAR-10, the training time for all methods are fairly similar. FedCKA extends training by the largest amount, as the matrix multiplication operation to calculate the CKA similarity is proportionally expensive to the forward and back propagation of the small model. 
However, for ResNet-50 trained on Tiny ImageNet, we see that the training time of FedProx, SCAFFOLD, and MOON have increased exponentially. Only FedCKA has comparable training times to FedAvg. 
This is because FedProx and SCAFFOLD performs expensive operations on the weights of each layer, and MOON performs forward propagation on three models until the penultimate layer. All these operation scale exponentially as the number of layers increase.
While FedCKA also performs forward propagation on three models, the number of layers remains static, thus being most efficient with medium sized models.

We emphasize that regularization must remain scalable for Federated Learning to be applied to state-of-the-art models. Even on ResNet-50, which is no longer considered a large model, other Federated Learning regularization methods lack scalabililty. This causes difficulty to test these methods with the current state-of-the-art models such as ViT \cite{dosovitskiy_vit} having 1.843 billion parameters, or slightly older models such as EfficientNet-B7 \cite{tan_effnet} having 813 layers.

\section{Conclusion and Future Work}
Improving the performance of Federated Learning on heterogeneous data is a widely researched topic. However, many previous works have incorrectly suggested that regularizing every layer of neural networks during local training is the best method to increase performance. 
We propose FedCKA, built on the most up-to-date understanding of neural networks.
By regularizing naturally similar, but not naturally dissimilar layers during local training, performance improves beyond previous works.
We also show that FedCKA is the only existing regularization method with adequate scalability when trained with a moderate sized model.

FedCKA shows that the proper regularization of important layers improves the performance of Federated Learning on heterogeneous data. However, standardizing the comparison of neural networks is an important step in a deeper understanding of neural networks. Moreover, there are questions as to the accuracy of CKA in measuring similarity in models such as Transformers or Graph Neural Networks. These are some topics we leave for future works.


\bibliography{aaai22}

\end{document}